\documentclass[a4paper,11pt,notitlepage]{article}
\usepackage[top=2cm,bottom=2cm,left=2cm,right=2cm,includefoot]{geometry}
\usepackage[margin=10pt,font=small,
            labelfont=bf,
            labelsep=quad,figurename=Figure.,
	    tablename=Table.]{caption}
\usepackage{multirow}

\usepackage{ccaption}

\usepackage{setspace}
\usepackage{indentfirst}
\usepackage{array}
\usepackage{threeparttable}

\usepackage{color, colortbl}
\definecolor{Sandy}{rgb}{0.93, 0.92, 0.88}
\definecolor{LightCyan}{rgb}{0.88,1,1}
\definecolor{Gray}{gray}{0.9}

\usepackage{bm}
\usepackage{amsmath}
\usepackage{amsbsy}
\usepackage{amstext}
\usepackage{amsfonts}
\usepackage{amssymb}

\usepackage[numbers,super,sort&compress]{natbib}
\usepackage{mciteplus}

\usepackage{graphicx}

\usepackage{hyperref}
\hypersetup{colorlinks,citecolor=red,linkcolor=blue}

\usepackage[pagewise]{lineno}

\usepackage[T1]{fontenc} 

\usepackage{times} 
\usepackage{mathptmx} 

\usepackage{multirow} 

\usepackage[version=4]{mhchem} 

\usepackage{xr}

\usepackage{authblk} 

\makeatletter
\renewcommand\@biblabel[1]{#1.}
\makeatother

\usepackage[ruled]{algorithm2e}

\begin{document}

\title{Language models as master equation solvers}
\author[a]{Chuanbo Liu}
\author[b, *]{Jin Wang}
\affil[a]{State Key Laboratory of Electroanalytical Chemistry, Changchun Institute of Applied Chemistry, Chinese Academy of Sciences, Changchun, Jilin, P.R. China, 130022.}
\affil[b]{Department of Chemistry and of Physics and Astronomy, State University of New York, Stony Brook, New York, USA, 11794-3400.}
\date{* jin.wang.1@stonybrook.edu}

\maketitle

\begin{abstract}
	\textbf{
		Master equations are of fundamental importance in modeling 
		stochastic dynamical systems.
		However, solving master equations is challenging due to the exponential increase in the number of possible states or trajectories with the dimension of the state space. 
		In this study, we propose repurposing language models as a machine learning approach to solve master equations. 
		We design a prompt-based neural network to map rate parameters, initial conditions, and time values directly to the state joint probability distribution that exactly matches the input contexts. 
		In this way, we approximate the solution of the master equation in its most general form. 
		We train the network using the policy gradient algorithm within the reinforcement learning framework, with feedback rewards provided by a set of variational autoregressive models. 
		By applying this approach to representative examples, we observe high accuracy for both multi-module and high-dimensional systems. 
		The trained network also exhibits extrapolating ability, extending its predictability to unseen data. 
		Our findings establish the connection between language models and master equations, highlighting the possibility of using a single pretrained large model to solve any master equation.
	}

\end{abstract}

\section*{Main}
For open complex systems, energies, materials, and information are constantly exchanged with the connected external environment. 
Constrained by information conservation, the time evolution of the system is captured by a system of ordinary differential equations called a stochastic master equation, which defines the time evolution of the probability of the system's state \cite{gardinerHandbookStochasticMethods2004, weberMasterEquationsTheory2017}. 
The master equation can be viewed as a more general probabilistic form of the deterministic equations, with the latter being the small-noise limit of the former \cite{wangLandscapeFluxTheory2015, xuUnifyingDeterministicStochastic2021}.
However, solving the master equations is both analytically impossible and computationally formidable for large systems since the number of possible states increases exponentially with the dimensions of the state space, resulting in exponentially increased memory and computational costs.

A great deal of effort has been devoted to solving the master equations. 
Two main categories of methods are commonly employed: sample-based methods
\cite{gillespieGeneralMethodNumerically1976, zuckermanWeightedEnsembleSimulation2017, gillespieGuidedProposalsEfficient2019, carlsonQuantumMonteCarlo2015} and state-based methods
\cite{munskyFiniteStateProjection2006, caoAccurateChemicalMaster2016, caoStateSpaceTruncation2016}.
More recently, neural networks had also been proposed for directly representing the solutions of the master equation
\cite{carleoSolvingQuantumManybody2017, davisUseMixtureDensity2020, sukys_approximating_2022_iscience, wuSolvingStatisticalMechanics2019,tangNeuralnetworkSolutionsStochastic2023} or approximating the transition matrix \cite{jiangNeuralNetworkAided2021a}. 
By harnessing the universal approximation property of artificial neural networks, data-driven training presents advantages in terms of accuracy and computational speed compared to classical methods. 
However, construction of simulation-based training datasets are time-consuming hence limited the solution to a few statistics. 
For variational-based method, the joint probability can be accurately approximated, but only for time-specific state. 
Numerous model states have to be saved for downstream applications such as parameter screening. 
Furthermore, additional interpolation methods have to be adapted to represent solutions for time or rate parameters that are not included in the training process. 

Transformer-based large language models that have been pretrained on web-scale datasets have demonstrated a strong ability to generate text. 
These models show zero-shot and few-shot generalization across various natural language understanding tasks \cite{NEURIPS2020_1457c0d6, touvronLLaMAOpenEfficient2023, bommasaniOpportunitiesRisksFoundation2022, hendyHowGoodAre2023, chenHowRobustGPT32023, koconChatGPTJackAll2023, qinChatGPTGeneralPurposeNatural2023, yangExploringLimitsChatGPT2023, yeComprehensiveCapabilityAnalysis2023}. 
New capabilities can be unlocked by providing specific text prompts to shape the behavior of the pre-trained foundation model. 
This tuning process allows the model to handle new tasks and data distributions that were not encountered during training \cite{stiennonLearningSummarizeHuman2020a, abramsonImitatingInteractiveIntelligence2021, ouyangTrainingLanguageModels2022, christianoDeepReinforcementLearning2023, NEURIPS2022_9d560961}.
When valid prompts are provided, the language models generate words whose conditional joint distributions align with the training data given the provided context. 
We want to highlight that a direct correspondence can be established between natural language and stochastic dynamics. 
This can be accomplished by constructing a word space where the size of the vocabulary represents the range of states, and the length of the text corresponds to the dimension of the state space. 
Therefore, we believe that language models also have the potential to represent the conditional joint probability of stochastic dynamical systems, given prompts that contain information about the master equation.

In this work, we introduce the Master Equation Transformer (MET), which generates the conditional state joint probability as a solution to a well-defined master equation. The information necessary for solving the equation is encoded in the prompts. 
We train the model using reinforcement learning techniques and a set of lightweight reward models. 
This training process is self-supervised and does not require simulation algorithms, unlike the supervised methods that rely on stochastic simulation algorithms \cite{davisUseMixtureDensity2020, sukys_approximating_2022_iscience}. 
Through testing on representative stochastic dynamical systems, we demonstrate that MET accurately represents the state joint probability. Additionally, when provided with prompts, the trained model can effectively map rate parameters, initial states, and times directly to state joint distributions in parallel. As a result, trajectory ensembles can be generated. 
This stands in stark contrast to the neural-network chemical master equation (NNCME) method \cite{tangNeuralnetworkSolutionsStochastic2023}, where the model state must be saved for every combination of rate parameters, initial states, and time points. 
Furthermore, our trained model is capable of extrapolating its predictability of the state joint distributions for rate parameters and time points that were not included in the training set. 
As the model size increases, we expect our methods be applied to master equations of large sizes that describe real-world systems. 
They can also be seamlessly applied to downstream tasks such as inferring model parameters, conducting sensitivity analysis on critical parameters, controlling stochastic systems based on the generated trajectory ensembles, and many more. 
Overall, we provide a universal approach for encoding solutions to the dynamics of stochastic systems.

\section*{Results}
We trained MET on representative examples of the chemical master equations (CME) that describe the stochastic dynamical behavior of chemical reaction networks. 
The dynamics of the genetic toggle switch \cite{gardnerConstructionGeneticToggle2000, terebusDiscreteContinuousModels2019} and mRNA turnover \cite{sukys_approximating_2022_iscience, caoComputationalModelingEukaryotic2001, suterMammalianGenesAre2011} were used to demonstrate that MET can represent master equation solutions with multimodal distributions and in high dimensions. 
We also trained MET on the autoregulatory feedback loop system \cite{sukys_approximating_2022_iscience} with varying rate parameters to show the potential of using MET to explore the parameter space. 
By using the trained network, efficient parameter screening can be conducted directly, and rate parameters can be inferred using a Markov-chain Monte Carlo method. 
Finally, we trained MET on the birth-death model with different initial states and generated the trajectory ensemble using the trained network.

To demonstrate the effectiveness of MET, we benchmark our results with those obtained using a simulation-based numerical method and a neural network-based method.
Specifically, we use the Gillespie simulation algorithm \cite{gillespieGeneralMethodNumerically1976} to simulate independent trajectories and estimate the marginal and joint statistics for all examples. 
Additionally, we generate state samples using NNCME method \cite{tangNeuralnetworkSolutionsStochastic2023} and compare them with the sampling states of MET.
For each model, we train a Recurrent Neural Network (RNN) according to the description provided in Tang et al. \cite{tangNeuralnetworkSolutionsStochastic2023} for rate parameters, initial states and time point values in accordance with the prompts provided to MET. 
We generate $10^4$ trajectories of samples for each system and each method, and display the marginal statistics and joint distributions to emphasize the accuracy of our method. 
We next provide the results for all these examples. 

\subsection*{Solving master equation with solution of multimodal distribution}
\begin{figure}
	\includegraphics[width=\linewidth]{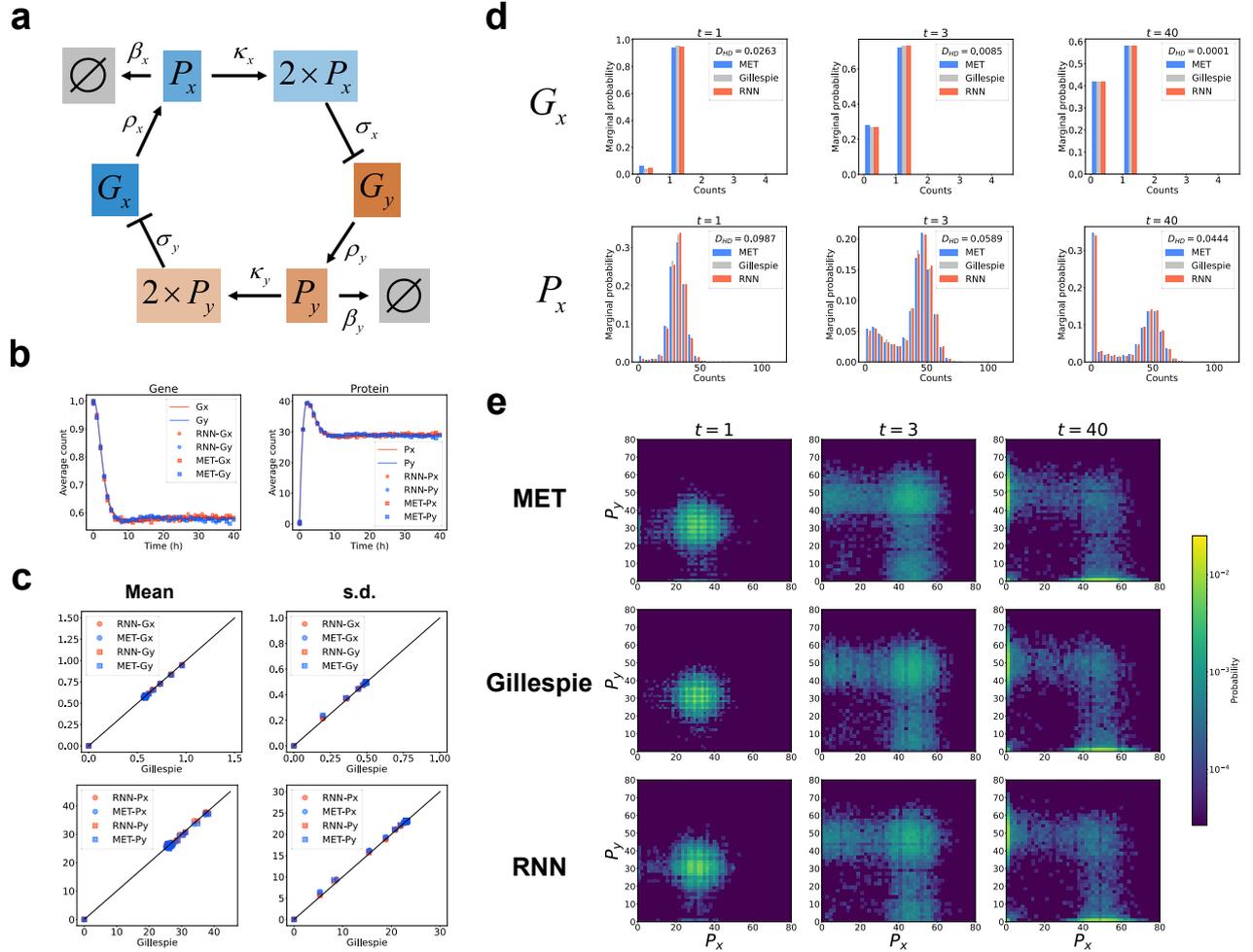}
	\medskip
	\caption{\textbf{Solving the time-dependent state joint distribution of the genetic toggle switch model.}
	\textbf{a}, Schematic of the genetic toggle switch model. 
	\textbf{b}, Average species counts as a function of time for the genes and proteins as compared with RNN and Gillespie results. 
	\textbf{c}, Comparison of mean and standard deviation for species counts of genes and proteins.  
	Results from RNN and MET are labeled with red and blue colors, respectively.
	\textbf{d}, The marginal distributions obtained from MET, RNN and Gillespie at time points $t=1$, $3$ and $40$. 
	The inset is the Hellinger distance between the distribution obtained from MET and Gillespie.
	\textbf{e}, The joint distributions of the two proteins from MET, RNN and Gillespie at time points $t=1$, $3$ and $40$.
	The color bar shows the joint probability values in the logarithmic scale. 
	$10^4$ states were sampled for MET and RNN at each time points, and Gillespie simulation has also $10^4$ trajectories. 
	}
	\label{fig:toggle}
\end{figure}

In this case study, we consider the genetic toggle switch system (see \autoref{fig:toggle} \textbf{a} and Supplementary Note).
The state space is a $6$-dimensional discrete space labeled by the number of two genes $G_x, G_y$, two proteins $P_x, P_y$ and the number of two protein homo-dimers $2\times P_x, 2\times P_y$. 
In this simplified reaction network, proteins express from the corresponding gene can form dimers that block the expression of the other gene. 
Under the chosen rate parameters, four modules can be identified from the joint probability $p(P_x, P_y)$.
MET was trained with a set of reward models consists of 100 different time points, and the rate parameters and initial states for constructing the prompts are invariant.
The trained MET correctly captures the time evolution statistics, including the mean and standard deviation (\autoref{fig:toggle} \textbf{b}) and matches the results sampling from Gillespie simulation (\autoref{fig:toggle} \textbf{c}). 
The comparison of marginal distributions at time $t=1, 3$ and $40$ on the state value of $G_x$ and $P_x$ are also consistent with both the simulation and neural network-based method (\autoref{fig:toggle} \textbf{d}).
The accuracy of the marginal distributions can be identified by the Hellinger distance \cite{hellingerNeueBegrundungTheorie1909} between results from MET and Gillespie.
The joint distribution $p(P_x, P_y)$ shows the identified four stable states of the two genes as compared to the results obtained from Gillespie simulation and state samples from RNN (\autoref{fig:toggle} \textbf{e}). 
These results are consistent with previous methods \cite{caoAccurateChemicalMaster2016, sukys_approximating_2022_iscience, tangNeuralnetworkSolutionsStochastic2023}. 
Note most of the sampled time points are not included in the training set, showing that the trained network has the power to extrapolate into unseen conditions.

\subsection*{Solving master equation with solution in high-dimensional state space}
We study the eukaryotic mRNA turnover system in this example.
The mRNA turnover system is a biological reaction network with a series of first-order reactions with a branch topology (see \autoref{fig:turnover} \textbf{a} and Supplementary Note). 
The gene state switching is also included in the reactions to account for transcriptional bursting \cite{suterMammalianGenesAre2011}.
The state space has a dimension of 17, and there are $25$ ordinary differential equations that describe the transitions between states. 
We trained the MET using a set of reward models corresponding to 100 different time points, keeping the rate parameters and initial states constant within the prompts (see Supplementary Note for more details).
By sampling from the trained MET, we found that the average counts of each species in the state space correctly match the time-evolution of the system (\autoref{fig:turnover} \textbf{b}).
The mean and standard deviation of the sampled states from MET consistently match the simulation trajectory samples (\autoref{fig:turnover} \textbf{c}).
The marginal and joint distributions shown in (\autoref{fig:turnover} \textbf{d}) demonstrates the pairwise comparison between the results obtained from MET, Gillespie or RNN at time $t=10$ (comparison at another time point $t=2$ is shown in the Supplementary Note).
These results are also consistent with the results obtained from previous study \cite{sukys_approximating_2022_iscience}.
Therefore, our results demonstrate that MET has the ability to accurately sample the joint probability of the system's state with a high-dimensional state space for every time point, including those excluded in the training set.

\subsection*{Exploring the parameter space}
\begin{figure}
	\includegraphics[width=\linewidth]{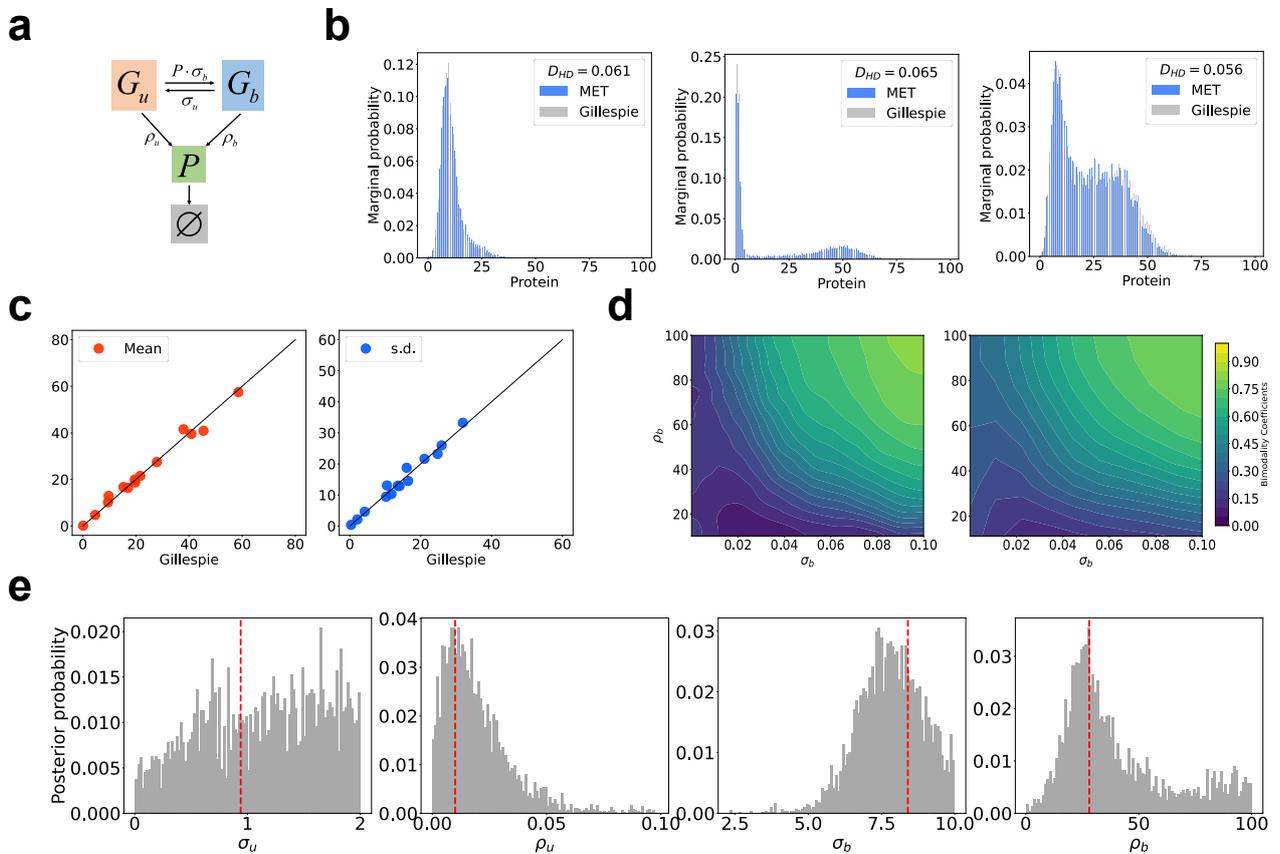}
	\medskip
	\caption{\textbf{Exploring the parameter space of the autoregulatory feedback loop model.}
	\textbf{a}, Schematic of the autoregulatory feedback loop model. 
	\textbf{b}, Marginal distribution of protein counts at time $t=10$ for 3 different parameter combinations (see Supplementary Notes for more details). 
	\textbf{c}, Comparison of the mean and standard deviations of 14 parameters sampled from MET and Gillespie.
	\textbf{d}, Prediction of the bimodality coefficients from samples of trained MET as compared to the groundtruth (right figure).
	For MET, $10^4$ pairs of $\sigma_b, \rho_b$ was sampled with each pair $10^3$ sampled states. 
	For Gillespie, $10^3$ trajectories was generated for 100 parameter points. 
	\textbf{e}, Inferencing model rate parameters with the trained MET network. Dashed vertical lines indicate the correct values of the tested case.
	The histograms were the results of $10^4$ Monte Carlo steps.
	}
	\label{fig:autoreg}
\end{figure}

After observing MET's ability to extrapolate its predictability to unseen conditions, we wonder if it can also be used to explore the parameter space.
To test this possibility, we utilize an autoregulatory feedback loop model \cite{sukys_approximating_2022_iscience}  (see \autoref{fig:autoreg} \textbf{a} and Supplementary Note for details). 
Initially, we train MET using a set of reward models that correspond to 128 different combinations of rate parameters and 50 different time points for each parameter combination.
The initial state remains unchanged for all prompts used.
To evaluate the accuracy of the trained model, we randomly select 14 combinations of rate parameters to sample the marginal distribution and evaluate the mean and standard deviations of the states (see Supplementary Note for more details). 
The sampled results from MET are then compared to those obtained from Gillespie's algorithm, and the distribution distances are calculated using the Hellinger distance between the two.
The results, presented in \autoref{fig:autoreg} \textbf{b} and \textbf{c} show that the means, standard deviations, and marginal distributions of counts align with the Gillespie sampling results.

To demonstrate MET's capability to explore the parameter space, we fix two rate parameters $\sigma_u = 1$ and $\rho_u = 1$, and calculate the bimodality coefficients for $\rho_b$ and $\sigma_b$ within the parameter region (see \autoref{fig:autoreg} \textbf{a} and Supplementary Note for more details).
In \autoref{fig:autoreg} \textbf{d}, we compare the results at $t=10$ with the ground truth values calculated from Gillespie's samplings.
Despite being trained on only 128 points in the continuous parameter space, MET accurately predicts the bimodality coefficients for up to $10^4$ combinations of rate parameters.
Finally, we test MET's performance in a parameter inference task.
We generate $10^4$ trajectories using a simulation algorithm and select 10 time points evenly as the experimental data.
We employ a Markov-chain Monte Carlo-based inference algorithm along with MET to infer the original rate parameters.
For each Monte Carlo step, the rate parameters undergo a random walk following a Gaussian kernel.
At each step, $10^3$ state samples are randomly chosen from the experimental data by first selecting the time points. 
From these samples, $2 \times 10^3$ prompts are constructed based on the chosen states and the rate parameters before and after the step. 
MET evaluates the likelihoods of the chosen states for both the original and new rate parameter combinations. 
The movement from the original position to the new position is only allowed if the average probability for the new rate parameter combination is greater than the original.
This algorithm is a Markov-chain based maximum likelihood estimation which will converge after sufficient large number of steps. 
We collect the visited rate parameters, and their histograms are shown in \autoref{fig:autoreg} \textbf{e}. 
Despite the limited number of rate parameter points in the training set, the inferencing algorithm provides accurate estimations for $\rho_u$, $\sigma_b$ and $\rho_b$. 
However, it fails to infer the value of $\sigma_u$, as we have restricted $\sigma_u$ to a small value region, which has a weak influence on the state distributions.
We anticipate that adding more training points to the reward model set will yield improved results.

\subsection*{Sampling of trajectory ensemble}
\begin{figure}
	\includegraphics[width=\linewidth]{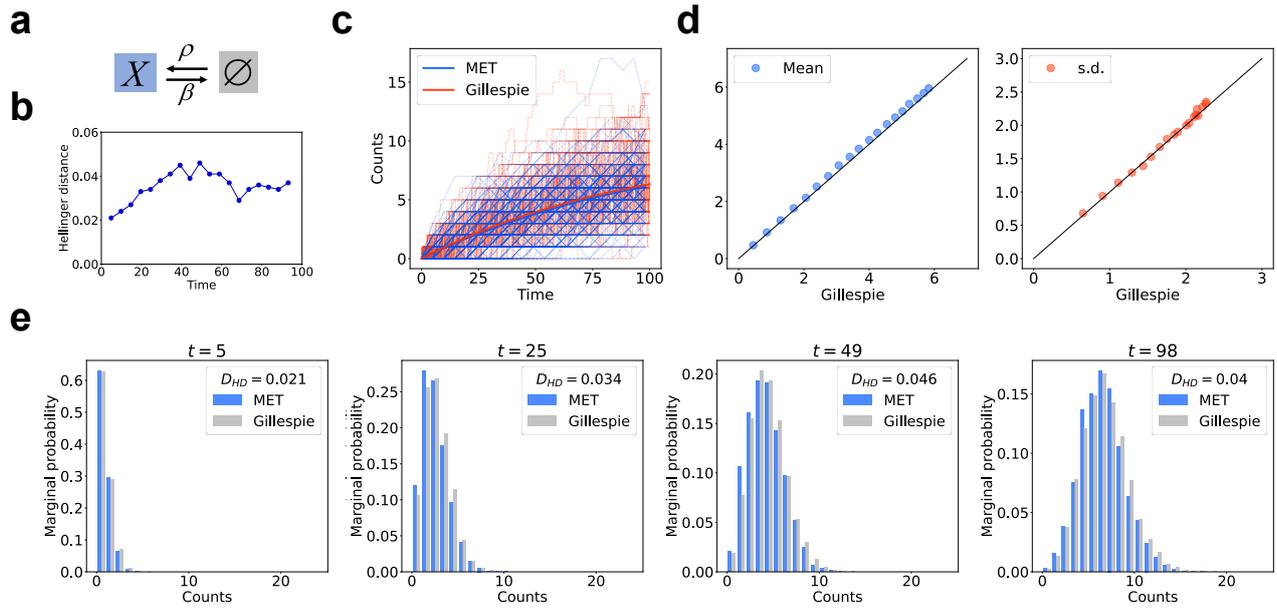}
	\medskip
	\caption{\textbf{Trajectory ensemble sampling of the birth-death model.}
	\textbf{a}, Schematic of the birth-death model. 
	\textbf{b}, Hellinger distance between the marginal distributions of trajectories sampled from MET and trajectories sampled by Gillespie for each time point. 
	\textbf{c}, Time-dependent trajectories sampled from MET (blue) and Gillespie (red). 
	The first 300 trajectories are used for plotting. 
	Average counts are shown as solid lines with respective colors. 
	\textbf{d}, Mean and standard deviations of time-specific species counts for trajectories sampled from MET as compared with Gillespie's results. 
	\textbf{e}, The marginal count distributions at time points $t = 5$, $25$, $49$ and $98$. The inset contains the corresponding Hellinger distance between the distributions sampled from MET and Gillespie. 
	}
	\label{fig:bd}
\end{figure}

For the last example, we consider the time-homogeneous birth-death process \cite{gardinerHandbookStochasticMethods2004} (\autoref{fig:bd} \textbf{a}).
We used a set of reward models corresponding to 20 different initial states, and each initial state contains 5 different time points for training MET.
The range of the time points is within $[0, 5]$.
The rate parameters remain unchanged for all the constructed prompts.
We then used the trained MET to sample trajectory ensembles.
Specifically, for each sampled trajectory, we construct prompts with fixed time values, and for each sampling step, the state value of the last step is used as the initial state value in the current step.
This way, by constructing the prompts iteratively, trajectories can be sampled in parallel (\autoref{fig:bd} \textbf{c}). 
We observe that the sampled trajectory ensemble effectively captures the marginal distributions throughout the time range (\autoref{fig:bd} \textbf{b}, \textbf{e}) as indicated by the Hellinger distance between the sampled trajectories from MET and Gillespie for $t=5, 25, 49$ and $98$. 
The mean and standard deviations of the state counts calculated from the sampled MET trajectories match those of the trajectory ensemble sampled from Gillespie (\autoref{fig:bd} \textbf{d}).
Although MET was trained only within the time region $[0, 5]$, we observe that it has the ability to gradually sample trajectories up to $t = 100$ to reach the steady-state.

\section*{Discussion}
The general solution of a well-defined master equation, which represents the conditional joint probability of the state, should include the dynamical system information as its contextual message. 
However, currently available methods only focus on approximating the solution for a specific time with fixed parameters and initial states.
In this study, we establish a connection between language models and the solution of master equations through the introduction of the MET network.
After training, MET learns a direct mapping from the provided prompts, which encode the dynamical system information, to the joint probability function of the state.
With MET, the knowledge of the stochastic dynamical system described by a master equation is compressed into a single model. 
The prompts serve as query keywords for extracting the solutions of the master equation under specified conditions. 
By providing valid prompts, MET can calculate the joint probability of state samples for any combination of rate parameters, initial states, and time values.
Furthermore, MET can also generate state samples with joint probabilities that accurately approximate the solution of the master equation by using prompts that include the system information. 
In contrast to previous methods, MET does not track the solution but simply provides it effortlessly. 
The computational cost of sampling is linear, rather than exponential, with respect to the dimension of the state space \cite{tayEfficientTransformersSurvey2022}.
Additionally, the computation of joint probabilities and the generation of state samples are performed in parallel, which greatly facilitates downstream tasks such as system state prediction, parameter inference, stochastic control, and so on.

By representing the solution of the master equation using language models, the methodology and models presented in the field of natural language processing can be smoothly applied to dynamical systems. 
The vocabulary size used by GPT is 50,257 during its training, and GPT-4 is capable of handling 32,768 tokens \cite{openaiGPT4TechnicalReport2023}, which is roughly the same number as the proteins within a cell. 
Additionally, it has been demonstrated that the length of context can be extended to $10^{10}$ tokens \cite{dingLongNetScalingTransformers2023}. 
Consequently, we anticipate that our approach will enable the accurate solving of numerous large real-world systems.

Moreover, we have observed that the trained MET model exhibits the ability to extrapolate to rate parameters and time points that were not included in the training set. 
This suggests that the network has either learned the causal relationship between the input prompts and the joint probability distribution, or it has somehow interpolated a solution based on the examples in the training set. 
Given that language models of considerable size have displayed emergent behavior \cite{openaiGPT4TechnicalReport2023}, it would be intriguing to ascertain whether increasing the size of the MET network would also result in similar emergent abilities.


Lastly, our results also raise the question of whether it is possible to pretrain a large model with various combinations of parameters and different initial states to create a universal model for all master equations. 
Training this foundational model can be computationally expensive, but the good news is that it only needs to be trained once and for all.

\section*{Methods}

\subsection*{Stochastic master equations and their solutions}

A well-defined master equation consists of 5 basic elements ``GRIBC'': (G) a directed graph which defines the connection between states; (R) a set of rate parameters modulate the speed of probability transitions; (I) the initial state distribution; (B) the state boundary conditions; and (C) the possible controls exerted by out-system sources. 
In this paper, we do not deal with controls and assumes the state space is discrete and is connected by an invariant finite directed graph.
Additionally, we always assume that the initial state distribution is a $\delta$-function centered at a specified state and use initial state distribution with initial state interchangeably.
A representative case within our scope are the chemical master equation \cite{geChemicalMasterEquation2013}.

A master equation defined under ``GRIBC'' describes the time-evolution of the state joint probability function $p(\mathbf{x}), \mathbf{x} \in \mathcal{H}$ under information conservation constrains, where $\mathcal{H} \in \mathbb{R}^N$ is the state space with dimension $N$. 
$p(\mathbf{x})$ is then the measure $\mathcal{H} \mapsto \mathbb{R}$.
The master equation is a kind of Komogorov forward equation whose solution is the conditional distribution $p_t(\mathbf{x} | \mathbf{\sigma})$, provided the initial condition $\mathbf{x}_0$, $t_0$.
\begin{equation}\label{eq:me}
	\partial_t p_t(\mathbf{x} | \mathbf{\sigma}) = \sum_{i=1}^M \left[\mathbf{W}_i(\mathbf{x} - \mathbf{\nu}) p_t (\mathbf{x} - \mathbf{\nu}) - \mathbf{W}_i' (\mathbf{x} | \mathbf{\sigma}) p_t (\mathbf{x} | \mathbf{\sigma}) \right]
\end{equation}
where $\mathbf{\sigma}$ are the rate parameters that defines the system timescale, $\mathbf{W}$ is the probability transition rate matrix, $\mathbf{\nu}$ is the state movement vector determined by the connection between states. 
\autoref{eq:me} can be written compactly as 
\begin{equation*}\label{eq:me_compact}
	\partial_t \mathbf{P} = \mathbb{T}_{\mathbf{\sigma}, t} \mathbf{P}	
\end{equation*}
where generator matrix $\mathbb{T}$ is defined as 
\begin{equation*}
	\left\{
	\begin{aligned}
		T_{\xi \mu} = & W_{\xi \mu} \\
		T_{\xi \xi} = & -\sum_{\mu \neq \xi} W_{\mu \xi} \\
	\end{aligned}
	\right.
\end{equation*}
with $\xi$ and $\mu$ represent arbitrary states.
The trajectory ensemble distribution is given by the cumulative productions of the probabilities of different combinations of states $p(\mathbf{x}_{t_T}, \ldots, \mathbf{x}_{t_0} | \mathbf{\sigma})$ as shown in \autoref{fig:summary}.

The solution $p(\mathbf{x})$ represents the joint distribution of the system's state at a specific time point with invariant rate parameters and initial state distribution. 
The most general form of the solution for the ``GRIBC'' well-defined master equation has the form $p_t(\mathbf{x} | \mathbf{\sigma}, \mathbf{x}_0, t_0)$ which is a probability measure of the state space $\mathcal{H}$ and is also a function of the rate parameters $\mathbf{\sigma}$, initial state $\mathbf{x}_0$, and time value $t$. 
Numerically representing $p(\mathbf{x})$ requires storing a real number for every configuration of the system.
If the state space has maximum of $U$ discrete states for each dimension, then storing $p(\mathbf{x})$ costs $\mathcal{O}(U^N)$ bytes at one time point.
If $T$ time points are sampled, the memory space required to store the trajectory ensemble distribution $p(\mathbf{x}_{t_T}, \ldots, \mathbf{x}_{t_0})$ will cost $\mathcal{O}(U^{NT})$ bytes. 
Since the solution of master equation also depends on $\mathbf{\sigma}$, $\mathbf{x}_0$, representing the general solution requires even more memory space. 
The number of possible state jumps (which is the number of ordinary differential equations in the master equation) is on the order of $\mathcal{O}(N^{2})$, so the dimension of the parameter space is $V^{N^2}$ if $V$ rate values are considered for each transition.
Therefore, directly representing the general solution of the master equation $p_t(\mathbf{x} | \mathbf{\sigma}, \mathbf{x}_0, t_0)$ will require $\mathcal{O}(U^{NTU^NV^{N^2}})$ bytes of memory.
Finally, if continuous parameters and time are considered, the memory space needed to represent the entire dynamics of the system is infinite.

\subsection*{Architecture of the neural network}
Based on the variational approach to approximate the joint distribution of states given prompts that encode information from the master equation, we design MET with masked multi-head self-attention blocks as its core functional module which describes the connection between prompts and states.

For calculating the self-attention score, we use learnable embeddings to convert the state vector $\mathbf{x} \in \mathbb{R}^{N}$ to a tensor in the embedding space $X \in \mathbb{R}^{N \times d_{emb}}$.
On the other hand, we collect the logarithm of rate parameters, the initial state and the time value in sequence to construct the prompt.
To handle the continuous variables of rate parameters and times in the prompt, we first transform rate parameters into logarithmic values to reduce value divergences. 
Then, we employ a one-layer perceptron to project the prompt to a vector $\mathbf{p}$ with fixed length $d_p$.
The reason for using fixed length for $\mathbf{p}$ is to ensure consistent input for different models.
We then map $\mathbf{p}$ to the same embedding space as the discrete-valued states with linear projector, to obtain the tensor $Y \in \mathbb{R}^{d_p \times d_{emb}}$.
The concatenation of $X$ and $Y$ serves as the input tensor $Z = \text{Concat}(X, Y) \in \mathbb{R}^{d_s \times d_{emb}}$ to the network, where $d_s = N+d_p$.
Once the prompts and states are embedded into the same space and the two tensors are concatenated, positional encoding is applied to the concatenated inputs to make use of the order of the input sequence. 
The masked multi-head self-attention block calculates the scaled dot-product attention of the input tensor $Z$ with $h$ attention heads \cite{vaswaniAttentionAllYou2023}
\begin{equation*}
	\begin{aligned}
		\text{Attention}(Z) & = \text{Concat}(\text{head}_1, \ldots, \text{head}_h) W_O \\
		\text{where} \quad \text{head}_i & = \text{Softmax} \left(\frac{\text{Mask}(Q_iK_i^T)}{\sqrt{d_k}}\right) V_i
	\end{aligned}
\end{equation*}
where the query $Q_i = Z W_{i, Q} \in \mathbb{R}^{d_s \times d_k}$, the key $K_i = Z W_{i, K} \in \mathbb{R}^{d_s \times d_k}$, and the value $V = Z W_{i, V} \in \mathbb{R}^{d_s \times d_v}$ are all linear projections of the input tensor $Z$.
These projections are obtained using different learnable parameter matrices: $W_{i, Q} \in \mathbb{R}^{d_{emb} \times d_k}$, $W_{i, K} \in \mathbb{R}^{d_{emb} \times d_k}$, $W_{i, V} \in \mathbb{R}^{d_{emb} \times d_v}$, and $W_O \in \mathbb{R}^{hd_v \times d_{emb}}$. 
Here, $d_k$ and $d_v$ represent the dimensions of the number of query (key) elements and value elements for each state value, respectively.
When $h$ is specified, $d_k$ and $d_v$ are defined according to the relation $d_k = d_v = d_{emb} / h$.
The attention mask $\text{Mask}(\cdot)$ is used to conceal the dependency relations between the previous position and subsequent positions of both the prompt and the generated sample states.
This is achieved by assigning $(Q_i K_i^T)_{\xi < \mu} = - \infty$, where $\xi$ and $\mu$ are the row and column indexes of the $Q_i K_i^T$ matrix, respectively.

We stack multiple attention blocks to create a multi-layer decoder, which enhances network's representational capability.
The output of the masked multi-head attention block $A \in \mathbb{R}^{d_s \times d_{emb}}$ is used as the input for another block. 
Following the multi-layer decoder, a language model head is utilized to map the decoder outputs to probability weights, which are then normalized using the softmax function.
The outputs of the MET provide joint distributions of states given the input prompts, as shown in \autoref{fig:summary}.
The order in which the species are arranged does not affect the results and only determines how to interpret the output probabilities of the neural network \cite{tangNeuralnetworkSolutionsStochastic2023}. This model architecture is similar to the language model used by the Generative Pre-trained Transformer (GPT) \cite{NEURIPS2020_1457c0d6,openaiGPT4TechnicalReport2023}.

\subsection*{Training methodology}
For training large language models, reinforcement learning with human feedback (RLHF) is a powerful method that aligns pre-trained language models with complex human tendencies \cite{abramsonImitatingInteractiveIntelligence2021, christianoDeepReinforcementLearning2023, ouyangTrainingLanguageModels2022, stiennonLearningSummarizeHuman2020a, zieglerFineTuningLanguageModels2020}.
We employ a similar approach by replacing human feedback with model feedback, which we term reinforcement learning with model feedback (RLMF).
RLMF begins by training a set of lightweight reward models for the provided master equation under different combinations of rate parameters and initial states, as described in \cite{tangNeuralnetworkSolutionsStochastic2023}. 
The network state for each time point is saved and formed a model set for training MET. 
The time costs of training the reward models are approximately linear with respect to the state space dimension as well as the total time steps \cite{tangNeuralnetworkSolutionsStochastic2023}.
Once the reward models are trained, they can be used together with the master equation to provide reward feedback on any generated states.
One advantage of using RLMF is that the information of $p(\mathbf{x})$ is highly compacted within the reward model, so the size of the model set is small and only linearly dependent on the total steps of solving the master equation. 
Conversely, if the Gillespie method is used to generate sparse samples in the giant trajectory ensemble space, the size of the dataset will exponentially increase in order to fully cover all possible states and give rewards based on the generated states.

For training MET, we minimize the relative-entropy (KL-divergence) loss
\begin{equation}\label{eq:loss}
	\mathcal{L} = D_{KL} \left[\hat{p}_{\mathbf{\theta}}(\mathbf{x}_{t+\delta t} | \mathbf{\sigma}, \mathbf{x}_0, t_0, \mathbf{\theta}_{t+\delta t}) || \hat{\mathbb{T}}\tilde{p}_{\tilde{\mathbf{\theta}}}(\mathbf{x}_{t} | \tilde{\mathbf{\theta}_{t}})\right]
\end{equation}
where $\hat{p}_{\mathbf{\theta}}$ is the distribution function approximated by MET, $\tilde{p}_{\tilde{\mathbf{\theta}}}$ is the distribution given by the corresponding reward model, and $\hat{\mathbb{T}} = e^{\delta t \mathbb{T}} \approx (\mathbb{I} + \delta t \mathbb{T})$ denotes the probability transition kernel in the time interval $\delta t$.
The gradient is 
\begin{equation*}\label{eq:gradient}
	\nabla_\theta \mathcal{L} = D_{KL} \left[\hat{p}_{\mathbf{\theta}}(\mathbf{x}_{t+\delta t} | \mathbf{\theta}_{t+\delta t}) || \hat{\mathbb{T}}\tilde{p}_{\tilde{\mathbf{\theta}}}(\mathbf{x}_{t} | \tilde{\mathbf{\theta}_{t}})\right] \nabla_\theta \ln \hat{p}_{\mathbf{\theta}}(\mathbf{x}_{t+\delta t} | \mathbf{\theta}_{t+\delta t})
\end{equation*}
where prompts are omitted. 
During the training process, combination of rate parameters, initial states and time points are randomly generated within the scope of the model set to construct the prompts. 
Within a reinforcement learning framework, $\hat{p}_{\mathbf{\theta}}$ is served as the stochastic policy to generate sample of states by MET based on the input prompts and its current weights, the generated samples are then used to approximate the gradient for weight updating 
\begin{equation*}\label{eq:gradient_mean}
	\nabla_\theta \mathcal{L} = \mathbf{E}_{\mathbf{x}_{t+\delta t} \sim \hat{p}} \left\{ \left[ \ln \hat{p}_{\mathbf{\theta}}(\mathbf{x}_{t+\delta t} | \mathbf{\theta}_{t+\delta t}) - \ln \hat{\mathbb{T}} \tilde{p}_{\tilde{\mathbf{\theta}}}(\mathbf{x}_{t} | \tilde{\mathbf{\theta}_{t}}) \cdot \right] \nabla_\theta \ln \hat{p}_{\mathbf{\theta}}(\mathbf{x}_{t+\delta t} | \mathbf{\theta}_{t+\delta t}) \right\}
\end{equation*} 
The difference of logarithm of probabilities gives the reward for taking the current policy. 
When training is converged, the model learns the desire policy which is the joint probability distribution corresponding to the input prompts. 
Since the monotonic property of the KL-divergence, the entropy-loss defined in \autoref{eq:loss} gives the upper-bound of the model accuracy. 
See \autoref{fig:summary} for schematic description of the training procedure.

\subsection*{Hyperparameters for training}
For MET, the hyperparameters include the dimension of the embedding space, denoted as $d_{emb}$, the number of feed-forward neurons, denoted as $d_{ff}$, the number of decoder layers, denoted as $d_l$, and the number of attention heads, denoted as $h$. 
During the training procedure, the elements in the model set are randomly shuffled before being fed to MET, and a one-step update uses the accumulated gradient of $M_{acc}$ randomly chosen elements. 
For each element in the gradient accumulation step, $S_{batch}$ state samples are drawn from the MET model, all of which have the same prompt. 
The value of $S_{batch}$ should be large enough to capture the divergence between $\hat{p}_{\mathbf{\theta}}$ and $\hat{\mathbb{T}}\tilde{p}_{\tilde{\mathbf{\theta}}}$, and should also be balanced with $M_{acc}$ in order to expose more prompt-target interchangeable dependencies. 
Otherwise, the network will tend to learn a single time point joint distribution. 
The prompts are constructed in the following order: rate parameters, initial states, and time. 
Rate parameters are represented with their logarithm values to reduce value divergence. 
For the master equation with an $N$-dimensional state space and $M$ accessible types of jumps, the prompt has a length of $N+M+1$, while the state has a length of $N$. 
The MET network was trained using the AdamW optimizer \cite{loshchilovDecoupledWeightDecay2019}, and a warming-up procedure was used to modulate the learning rate during training. 
The learning rates used for tuning warming-up and decay of the learning rate are set based on the size of the model. 
Typically, for a network with $d_{emb} = 64$, $d_{ff} = 1024$, $d_l = 8$, and $h = 8$, the network has approximately 0.4 million parameters, and the learning rate is set to be between $6 \times 10^{-4}$ and $1 \times 10^{-3}$. 
The chosen values of hyperparameters for training reward models and the MET network in this work are listed in Table \ref{tab:reward} and Table \ref{tab:hyperparameters}, respectively. 
We note that the current network is limited by the GPU memory (Tesla-V100) we used. 
Employing a larger network is expected to yield more accurate results, increase the representative ability for learning larger systems, and be applicable to systems with more combinations of prompts.

\newpage

\section*{Data \& code availability}
A pytorch implementation of our algorithm will be available upon the acceptance of the manuscript.

\section*{Acknowledgments}
C.-B.L. thanks the supports from the National Natural Science Foundation of China Grant 32000888, the Scientific Instrument Developing Project of the Chinese Academy of Sciences Grant YJKYYQ20180038,
Jilin Province Science and Technology Development Plan Grant 20230101152JC.

\section*{Author contributions}
C.-B.L. implemented the MET package, performed the numerical experiments, analyzed the data and wrote the original draft.
J.W. provided critical suggestions, revised the manuscript, and supervised the study.

\section*{Competing interests}
The authors declare no competing interests.

\section*{Correspondence}
Correspondence and requests for materials should be addressed to Jin Wang.

\newpage


\begin{thebibliography}{99}
	\expandafter\ifx\csname url\endcsname\relax
	  \def\url#1{\texttt{#1}}\fi
	\expandafter\ifx\csname urlprefix\endcsname\relax\def\urlprefix{URL }\fi
	\providecommand{\bibinfo}[2]{#2}
	\providecommand{\eprint}[2][]{\url{#2}}
	
	\bibitem{gardinerHandbookStochasticMethods2004}
	\bibinfo{author}{Gardiner, C.~W.}
	\newblock \emph{\bibinfo{title}{Handbook of Stochastic Methods for Physics, Chemistry, and the Natural Sciences}}.
	\newblock Springer Series in Synergetics (\bibinfo{publisher}{Springer-Verlag}, \bibinfo{address}{Berlin ; New York}, \bibinfo{year}{2004}), \bibinfo{edition}{3rd ed} edn.
	
	\bibitem{weberMasterEquationsTheory2017}
	\bibinfo{author}{Weber, M.~F.} \& \bibinfo{author}{Frey, E.}
	\newblock \bibinfo{title}{Master equations and the theory of stochastic path integrals}.
	\newblock \emph{\bibinfo{journal}{Reports on Progress in Physics}} \textbf{\bibinfo{volume}{80}}, \bibinfo{pages}{046601} (\bibinfo{year}{2017}).
	
	\bibitem{wangLandscapeFluxTheory2015}
	\bibinfo{author}{Wang, J.}
	\newblock \bibinfo{title}{Landscape and flux theory of non-equilibrium dynamical systems with application to biology}.
	\newblock \emph{\bibinfo{journal}{Advances in Physics}} \textbf{\bibinfo{volume}{64}}, \bibinfo{pages}{1--137} (\bibinfo{year}{2015}).
	
	\bibitem{xuUnifyingDeterministicStochastic2021}
	\bibinfo{author}{Xu, L.}, \bibinfo{author}{Patterson, D.}, \bibinfo{author}{Staver, A.~C.}, \bibinfo{author}{Levin, S.~A.} \& \bibinfo{author}{Wang, J.}
	\newblock \bibinfo{title}{Unifying deterministic and stochastic ecological dynamics via a landscape-flux approach}.
	\newblock \emph{\bibinfo{journal}{Proceedings of the National Academy of Sciences}} \textbf{\bibinfo{volume}{118}}, \bibinfo{pages}{e2103779118} (\bibinfo{year}{2021}).
	
	\bibitem{gillespieGeneralMethodNumerically1976}
	\bibinfo{author}{Gillespie, D.~T.}
	\newblock \bibinfo{title}{A general method for numerically simulating the stochastic time evolution of coupled chemical reactions}.
	\newblock \emph{\bibinfo{journal}{Journal of Computational Physics}} \textbf{\bibinfo{volume}{22}}, \bibinfo{pages}{403--434} (\bibinfo{year}{1976}).
	
	\bibitem{zuckermanWeightedEnsembleSimulation2017}
	\bibinfo{author}{Zuckerman, D.~M.} \& \bibinfo{author}{Chong, L.~T.}
	\newblock \bibinfo{title}{Weighted ensemble simulation: Review of methodology, applications, and software}.
	\newblock \emph{\bibinfo{journal}{Annual Review of Biophysics}} \textbf{\bibinfo{volume}{46}}, \bibinfo{pages}{43--57} (\bibinfo{year}{2017}).
	
	\bibitem{gillespieGuidedProposalsEfficient2019}
	\bibinfo{author}{Gillespie, C.~S.} \& \bibinfo{author}{Golightly, A.}
	\newblock \bibinfo{title}{Guided proposals for efficient weighted stochastic simulation}.
	\newblock \emph{\bibinfo{journal}{The Journal of Chemical Physics}} \textbf{\bibinfo{volume}{150}}, \bibinfo{pages}{224103} (\bibinfo{year}{2019}).
	
	\bibitem{carlsonQuantumMonteCarlo2015}
	\bibinfo{author}{Carlson, J.} \emph{et~al.}
	\newblock \bibinfo{title}{Quantum monte carlo methods for nuclear physics}.
	\newblock \emph{\bibinfo{journal}{Reviews of Modern Physics}} \textbf{\bibinfo{volume}{87}}, \bibinfo{pages}{1067--1118} (\bibinfo{year}{2015}).
	
	\bibitem{munskyFiniteStateProjection2006}
	\bibinfo{author}{Munsky, B.} \& \bibinfo{author}{Khammash, M.}
	\newblock \bibinfo{title}{The finite state projection algorithm for the solution of the chemical master equation}.
	\newblock \emph{\bibinfo{journal}{The Journal of Chemical Physics}} \textbf{\bibinfo{volume}{124}}, \bibinfo{pages}{044104} (\bibinfo{year}{2006}).
	
	\bibitem{caoAccurateChemicalMaster2016}
	\bibinfo{author}{Cao, Y.}, \bibinfo{author}{Terebus, A.} \& \bibinfo{author}{Liang, J.}
	\newblock \bibinfo{title}{Accurate chemical master equation solution using multi-finite buffers}.
	\newblock \emph{\bibinfo{journal}{Multiscale Modeling \& Simulation}} \textbf{\bibinfo{volume}{14}}, \bibinfo{pages}{923--963} (\bibinfo{year}{2016}).
	
	\bibitem{caoStateSpaceTruncation2016}
	\bibinfo{author}{Cao, Y.}, \bibinfo{author}{Terebus, A.} \& \bibinfo{author}{Liang, J.}
	\newblock \bibinfo{title}{State space truncation with quantified errors for accurate solutions to discrete chemical master equation}.
	\newblock \emph{\bibinfo{journal}{Bulletin of Mathematical Biology}} \textbf{\bibinfo{volume}{78}}, \bibinfo{pages}{617--661} (\bibinfo{year}{2016}).
	
	\bibitem{carleoSolvingQuantumManybody2017}
	\bibinfo{author}{Carleo, G.} \& \bibinfo{author}{Troyer, M.}
	\newblock \bibinfo{title}{Solving the quantum many-body problem with artificial neural networks}.
	\newblock \emph{\bibinfo{journal}{Science}} \textbf{\bibinfo{volume}{355}}, \bibinfo{pages}{602--606} (\bibinfo{year}{2017}).
	
	\bibitem{davisUseMixtureDensity2020}
	\bibinfo{author}{Davis, C.~N.}, \bibinfo{author}{Hollingsworth, T.~D.}, \bibinfo{author}{Caudron, Q.} \& \bibinfo{author}{Irvine, M.~A.}
	\newblock \bibinfo{title}{The use of mixture density networks in the emulation of complex epidemiological individual-based models}.
	\newblock \emph{\bibinfo{journal}{PLOS Computational Biology}} \textbf{\bibinfo{volume}{16}}, \bibinfo{pages}{e1006869} (\bibinfo{year}{2020}).
	
	\bibitem{sukys_approximating_2022_iscience}
	\bibinfo{author}{Sukys, A.}, \bibinfo{author}{{\"O}cal, K.} \& \bibinfo{author}{Grima, R.}
	\newblock \bibinfo{title}{Approximating solutions of the chemical master equation using neural networks}.
	\newblock \emph{\bibinfo{journal}{iScience}} \textbf{\bibinfo{volume}{25}}, \bibinfo{pages}{105010} (\bibinfo{year}{2022}).
	
	\bibitem{wuSolvingStatisticalMechanics2019}
	\bibinfo{author}{Wu, D.}, \bibinfo{author}{Wang, L.} \& \bibinfo{author}{Zhang, P.}
	\newblock \bibinfo{title}{Solving statistical mechanics using variational autoregressive networks}.
	\newblock \emph{\bibinfo{journal}{Physical Review Letters}} \textbf{\bibinfo{volume}{122}}, \bibinfo{pages}{080602} (\bibinfo{year}{2019}).
	
	\bibitem{tangNeuralnetworkSolutionsStochastic2023}
	\bibinfo{author}{Tang, Y.}, \bibinfo{author}{Weng, J.} \& \bibinfo{author}{Zhang, P.}
	\newblock \bibinfo{title}{Neural-network solutions to stochastic reaction networks}.
	\newblock \emph{\bibinfo{journal}{Nature Machine Intelligence}} \textbf{\bibinfo{volume}{5}}, \bibinfo{pages}{376--385} (\bibinfo{year}{2023}).
	
	\bibitem{jiangNeuralNetworkAided2021a}
	\bibinfo{author}{Jiang, Q.} \emph{et~al.}
	\newblock \bibinfo{title}{Neural network aided approximation and parameter inference of non-markovian models of gene expression}.
	\newblock \emph{\bibinfo{journal}{Nature Communications}} \textbf{\bibinfo{volume}{12}}, \bibinfo{pages}{2618} (\bibinfo{year}{2021}).
	
	\bibitem{NEURIPS2020_1457c0d6}
	\bibinfo{author}{Brown, T.} \emph{et~al.}
	\newblock \bibinfo{title}{Language models are few-shot learners}.
	\newblock In \bibinfo{editor}{Larochelle, H.}, \bibinfo{editor}{Ranzato, M.}, \bibinfo{editor}{Hadsell, R.}, \bibinfo{editor}{Balcan, M.} \& \bibinfo{editor}{Lin, H.} (eds.) \emph{\bibinfo{booktitle}{Advances in Neural Information Processing Systems}}, vol.~\bibinfo{volume}{33}, \bibinfo{pages}{1877--1901} (\bibinfo{publisher}{Curran Associates, Inc.}, \bibinfo{year}{2020}).
	
	\bibitem{touvronLLaMAOpenEfficient2023}
	\bibinfo{author}{Touvron, H.} \emph{et~al.}
	\newblock \bibinfo{title}{Llama: Open and efficient foundation language models}. \newblock \bibinfo{eprint}{Preprint at https://arxiv.org/abs/2302.13971} (\bibinfo{year}{2023}).
	
	
	\bibitem{bommasaniOpportunitiesRisksFoundation2022}
	\bibinfo{author}{Bommasani, R.} \emph{et~al.}
	\newblock \bibinfo{title}{On the opportunities and risks of foundation models}. \newblock \bibinfo{eprint}{Preprint at https://arxiv.org/abs/2108.07258} (\bibinfo{year}{2022}).
	
	\bibitem{hendyHowGoodAre2023}
	\bibinfo{author}{Hendy, A.} \emph{et~al.}
	\newblock \bibinfo{title}{How good are GPT models at machine translation? A comprehensive evaluation}. \newblock \bibinfo{eprint}{Preprint at https://arxiv.org/abs/2302.09210} (\bibinfo{year}{2023}).
	
	\bibitem{chenHowRobustGPT32023}
	\bibinfo{author}{Chen, X.} \emph{et~al.}
	\newblock \bibinfo{title}{How robust is GPT-3.5 to predecessors? a comprehensive study on language understanding tasks}. \newblock \bibinfo{eprint}{Preprint at https://arxiv.org/abs/2303.00293} (\bibinfo{year}{2023}).
	
	\bibitem{koconChatGPTJackAll2023}
	\bibinfo{author}{Koco{\'n}, J.} \emph{et~al.}
	\newblock \bibinfo{title}{ChatGPT: Jack of all trades, master of none}.
	\newblock \emph{\bibinfo{journal}{Information Fusion}} \textbf{\bibinfo{volume}{99}}, \bibinfo{pages}{101861} (\bibinfo{year}{2023}).
	
	\bibitem{qinChatGPTGeneralPurposeNatural2023}
	\bibinfo{author}{Qin, C.} \emph{et~al.}
	\newblock \bibinfo{title}{Is chatGPT a general-purpose natural language processing task solver?}. \newblock \bibinfo{eprint}{Preprint at https://arxiv.org/abs/2302.06476} (\bibinfo{year}{2023}).
	
	\bibitem{yangExploringLimitsChatGPT2023}
	\bibinfo{author}{Yang, X.}, \bibinfo{author}{Li, Y.}, \bibinfo{author}{Zhang, X.}, \bibinfo{author}{Chen, H.} \& \bibinfo{author}{Cheng, W.}
	\newblock \bibinfo{title}{Exploring the limits of chatGPT for query or aspect-based text summarization}. \newblock \bibinfo{eprint}{Preprint at https://arxiv.org/abs/2302.08081} (\bibinfo{year}{2023}).
	
	\bibitem{yeComprehensiveCapabilityAnalysis2023}
	\bibinfo{author}{Ye, J.} \emph{et~al.}
	\newblock \bibinfo{title}{A comprehensive capability analysis of GPT-3 and GPT-3.5 series models}. \newblock \bibinfo{eprint}{Preprint at https://arxiv.org/abs/2303.10420} (\bibinfo{year}{2023}).
	
	\bibitem{stiennonLearningSummarizeHuman2020a}
	\bibinfo{author}{Stiennon, N.} \emph{et~al.}
	\newblock \bibinfo{title}{Learning to summarize from human feedback}.
	\newblock In \emph{\bibinfo{booktitle}{Proceedings of the 34th International Conference on Neural Information Processing Systems}}, NIPS'20 (\bibinfo{publisher}{Curran Associates Inc.}, \bibinfo{address}{Red Hook, NY, USA}, \bibinfo{year}{2020}).
	
	\bibitem{abramsonImitatingInteractiveIntelligence2021}
	\bibinfo{author}{Abramson, J.} \emph{et~al.}
	\newblock \bibinfo{title}{Imitating interactive intelligence}. \newblock \bibinfo{eprint}{Preprint at https://arxiv.org/abs/2012.05672} (\bibinfo{year}{2021}).
	
	\bibitem{ouyangTrainingLanguageModels2022}
	\bibinfo{author}{Ouyang, L.} \emph{et~al.}
	\newblock \bibinfo{title}{Training language models to follow instructions with human feedback}. \newblock \bibinfo{eprint}{Preprint at https://arxiv.org/abs/2203.02155} (\bibinfo{year}{2022}).
	
	\bibitem{christianoDeepReinforcementLearning2023}
	\bibinfo{author}{Christiano, P.} \emph{et~al.}
	\newblock \bibinfo{title}{Deep reinforcement learning from human preferences}. \newblock \bibinfo{eprint}{Preprint at https://arxiv.org/abs/1706.03741} (\bibinfo{year}{2023}).
	
	\bibitem{NEURIPS2022_9d560961}
	\bibinfo{author}{Wei, J.} \emph{et~al.}
	\newblock \bibinfo{title}{Chain-of-thought prompting elicits reasoning in large language models}.
	\newblock In \bibinfo{editor}{Koyejo, S.} \emph{et~al.} (eds.) \emph{\bibinfo{booktitle}{Advances in Neural Information Processing Systems}}, vol.~\bibinfo{volume}{35}, \bibinfo{pages}{24824--24837} (\bibinfo{publisher}{Curran Associates, Inc.}, \bibinfo{year}{2022}).
	
	\bibitem{gardnerConstructionGeneticToggle2000}
	\bibinfo{author}{Gardner, T.~S.}, \bibinfo{author}{Cantor, C.~R.} \& \bibinfo{author}{Collins, J.~J.}
	\newblock \bibinfo{title}{Construction of a genetic toggle switch in escherichia coli}.
	\newblock \emph{\bibinfo{journal}{Nature}} \textbf{\bibinfo{volume}{403}}, \bibinfo{pages}{339--342} (\bibinfo{year}{2000}).
	
	\bibitem{terebusDiscreteContinuousModels2019}
	\bibinfo{author}{Terebus, A.}, \bibinfo{author}{Liu, C.} \& \bibinfo{author}{Liang, J.}
	\newblock \bibinfo{title}{Discrete and continuous models of probability flux of switching dynamics: Uncovering stochastic oscillations in a toggle-switch system}.
	\newblock \emph{\bibinfo{journal}{The Journal of Chemical Physics}} \textbf{\bibinfo{volume}{151}}, \bibinfo{pages}{185104} (\bibinfo{year}{2019}).
	
	\bibitem{caoComputationalModelingEukaryotic2001}
	\bibinfo{author}{Cao, D.} \& \bibinfo{author}{Parker, R.}
	\newblock \bibinfo{title}{Computational modeling of eukaryotic mrna turnover}.
	\newblock \emph{\bibinfo{journal}{RNA}} \textbf{\bibinfo{volume}{7}}, \bibinfo{pages}{1192--1212} (\bibinfo{year}{2001}).
	
	\bibitem{suterMammalianGenesAre2011}
	\bibinfo{author}{Suter, D.~M.} \emph{et~al.}
	\newblock \bibinfo{title}{Mammalian genes are transcribed with widely different bursting kinetics}.
	\newblock \emph{\bibinfo{journal}{Science}} \textbf{\bibinfo{volume}{332}}, \bibinfo{pages}{472--474} (\bibinfo{year}{2011}).
	
	\bibitem{hellingerNeueBegrundungTheorie1909}
	\bibinfo{author}{Hellinger, E.}
	\newblock \bibinfo{title}{Neue begr\"undung der theorie quadratischer formen von unendlichvielen ver\"anderlichen.}
	\newblock \emph{\bibinfo{journal}{Journal f\"ur die reine und angewandte Mathematik}} \textbf{\bibinfo{volume}{1909}}, \bibinfo{pages}{210--271} (\bibinfo{year}{1909}).
	
	\bibitem{tayEfficientTransformersSurvey2022}
	\bibinfo{author}{Tay, Y.}, \bibinfo{author}{Dehghani, M.}, \bibinfo{author}{Bahri, D.} \& \bibinfo{author}{Metzler, D.}
	\newblock \bibinfo{title}{Efficient transformers: A survey}. \newblock \bibinfo{eprint}{Preprint at https://arxiv.org/abs/2009.06732} (\bibinfo{year}{2022}).
	
	\bibitem{openaiGPT4TechnicalReport2023}
	\bibinfo{author}{OpenAI}.
	\newblock \bibinfo{title}{Gpt-4 technical report}. \newblock \bibinfo{eprint}{Preprint at https://arxiv.org/abs/2303.08774} (\bibinfo{year}{2023}).
	
	\bibitem{dingLongNetScalingTransformers2023}
	\bibinfo{author}{Ding, J.} \emph{et~al.}
	\newblock \bibinfo{title}{Longnet: Scaling transformers to 1,000,000,000 tokens}. \newblock \bibinfo{eprint}{Preprint at https://arxiv.org/abs/2307.02486} (\bibinfo{year}{2023}).
	
	\bibitem{geChemicalMasterEquation2013}
	\bibinfo{author}{Ge, H.} \& \bibinfo{author}{Qian, H.}
	\newblock \bibinfo{title}{Chemical master equation}.
	\newblock In \bibinfo{editor}{Dubitzky, W.}, \bibinfo{editor}{Wolkenhauer, O.}, \bibinfo{editor}{Cho, K.-H.} \& \bibinfo{editor}{Yokota, H.} (eds.) \emph{\bibinfo{booktitle}{Encyclopedia of Systems Biology}}, \bibinfo{pages}{396--399} (\bibinfo{publisher}{Springer New York}, \bibinfo{address}{New York, NY}, \bibinfo{year}{2013}).
	
	\bibitem{vaswaniAttentionAllYou2023}
	\bibinfo{author}{Vaswani, A.} \emph{et~al.}
	\newblock \bibinfo{title}{Attention is all you need}. \newblock \bibinfo{eprint}{Preprint at https://arxiv.org/abs/1706.03762} (\bibinfo{year}{2023}).
	
	\bibitem{zieglerFineTuningLanguageModels2020}
	\bibinfo{author}{Ziegler, D.~M.} \emph{et~al.}
	\newblock \bibinfo{title}{Fine-tuning language models from human preferences}. \newblock \bibinfo{eprint}{Preprint at https://arxiv.org/abs/1909.08593} (\bibinfo{year}{2020}).
	
	\bibitem{loshchilovDecoupledWeightDecay2019}
	\bibinfo{author}{Loshchilov, I.} \& \bibinfo{author}{Hutter, F.}
	\newblock \bibinfo{title}{Decoupled weight decay regularization}. \newblock \bibinfo{eprint}{Preprint at https://arxiv.org/abs/1711.05101} (\bibinfo{year}{2019}).
	
\end{thebibliography}

\newpage

\begin{figure}[t]
	\centering
	\includegraphics[width=\linewidth]{figure1.pdf}
	\medskip
	\caption{\emph{Caption on next page.}}
	\label{fig:summary}
\end{figure}

\begin{figure}[t!]
	\contcaption{
	\textbf{Solving stochastic master equation with language model.}
	\textbf{Top left}, an open complex system constantly communicates energy, materials and information with the environment, which can be modeled by the stochastic master equation. 
	The solution of the master equation is the joint distribution of states at a set of time points $\{\mathbf{t}_0, \mathbf{t}_1, \ldots, \mathbf{t}_T\}$, where $t_T$ is the ending time point for observation. 
	The number of time points can be either countable or uncountable. 
	In the latter case, the system is modeled as a continuous-time model. 
	\textbf{Top right}, the network architecture of MET is shown, with the prompt containing the query information. 
	The prompt goes through a one-layer perceptron before being embedded in the same embedding space as the states. 
	The complete input to the language model is the concatenation of the prompt embedding and the state embedding. 
	The output of the language model is the conditional probabilities given the input prompt, while the joint probability is the accumulation of the output conditional probabilities. 
	Details of the network architecture are described in the Method section.
	\textbf{Middle}, MET is trained through a reinforcement learning with model feedback strategy. 
	Firstly, a set of lightweight reward models is trained under the supervision of the master equation. 
	Then MET is trained by generating state samples in context with the input prompts. 
	The true probability $\hat{\mathbb{T}}\tilde{p}_{\tilde{\mathbf{\theta}}}(\mathbf{x}_{t} | \tilde{\mathbf{\theta}_{t}})$ is calculated by the master equation based on the state probabilities $\tilde{p}_{\tilde{\mathbf{\theta}}}(\mathbf{x}_{t} | \tilde{\mathbf{\theta}_{t}})$ provided by the reward model.
	This ground truth probability is then compared with the self-evaluated probability $\hat{p}_{\mathbf{\theta}}(\mathbf{x}_{t+\delta t} | \mathbf{\sigma}, \mathbf{x}_0, t_0, \mathbf{\theta}_{t+\delta t})$ by MET.
	Policy gradient is therefore calculated following the method of proximal policy optimization with the gradient $\nabla_\theta \mathcal{L}$. 
	\textbf{Below}, the bilingual property of the language models. 
	The same language model can learn both sentences in natural language and the stochastic dynamics governed by the stochastic master equations. 
	After training, the language model can directly provide the joint probability from the input prompts, which state the rate parameters, the initial conditions, and the time elapsed from $t_0$.
	}
\end{figure}

\begin{figure}
	\includegraphics[width=0.9\linewidth]{figure_turnover.pdf}
	\medskip
	\caption{\emph{Caption on next page.}}
	\label{fig:turnover}
\end{figure}

\begin{figure}[t!]
	\contcaption{
	\textbf{Solving the time-dependent state joint distribution of the mRNA turnover model.}
	\textbf{a}, Schematic of the mRNA turnover model. 
	\textbf{b}, Average species counts as a function of time for 8 species as compared with RNN and Gillespie results. 
	Sample means for RNN and MET are represented as colored circles and squares, respectively. 
	\textbf{c}, Comparisons of mean and standard deviation of species counts  obtained from MET and Gillespie for 8 different species.  
	\textbf{d}, Marginal (diagonal graphics) and joint distributions sampled from MET (lower triangle) and Gillespie (higher triangle) at $t = 10$. 
	The insets of the diagonal graphics are the Hellinger distance between the marginal distribution obtained from MET and Gillespie.
	The color bars shows the joint probability values in the logarithmic scale. 
	$10^4$ states were sampled for MET and RNN at each time points, and Gillespie simulation has also $10^4$ trajectories. 
	}
\end{figure}

\begin{table}[ht]
	\centering
	\begin{threeparttable}
		\caption{\textbf{Hyperparameters used in this work for training the reward models.}}
		\label{tab:reward}
		\begin{tabular}{c|cccccccccc}
			\hline
			\rowcolor{Sandy}
			& Model Type & Depth & Width & Batch Size & $\delta t$ & Epochs & $r_{learn}$ & $U$ & $t_T$ \\
			\hline
			Autoreg & RNN & 1 & 32 & 1000 & $2\times 10^{-3}$ & 100 & 0.001 & 100 & 10 \\
			\hline
			\rowcolor{Sandy}
			Birth-death & RNN & 1 & 32 & 1000 & $1\times 10^{-2}$ & 100 & 0.001 & 10 & 100 \\
			\hline
			Cascade & RNN & 1 & 32 & 1000 & $1\times 10^{-2}$ & 100 & 0.001 & 10 & 10 \\
			\hline
			\rowcolor{Sandy}
			Gene express & RNN & 1 & 128 & 1000 & $0.1$ & 100 & 0.001 & 100 & 3600 \\
			\hline
			mRNA turnover & RNN & 1 & 128 & 1000 & $1\times 10^{-4}$ & 500 & 0.001 & 200 & 500 \\
			\hline
			\rowcolor{Sandy}
			Toggle switch & RNN & 1 & 128 & 1000 & $5\times 10^{-3}$ & 200 & 0.0006 & 80 & 40 \\
			\hline
		\end{tabular}
		\begin{tablenotes}
			\small
			\item
			We used a one-hidden layer Gate Recurrent Unit (GRU) model to train all the reward models. The width of the model was chosen based on the number of species in the CME. The training batch size was set to 1,000, and 100-500 epochs were trained for each time step. We used a constant training rate, denoted as $r_{learn}$, which was optimized to prevent cost function oscillation while ensuring fast convergence. The updating time step, $\delta t$, for solving the CME determines the total number of steps before reaching the final time, $t_T$. The upper bound of the state, $U$, was determined from trial Gillespie simulations. All the reward models were trained using a single core GPU of a Tesla-V100 on cloud servers.
		\end{tablenotes}
	\end{threeparttable}
\end{table}

\begin{table}[ht]
	\centering
	\begin{threeparttable}
		\caption{\textbf{Models used in this work and the training times under the chosen hyperparameters}}
		\label{tab:hyperparameters}
		\begin{tabular}{c|ccccccccccc}
			\hline
			\rowcolor{Sandy}
			& $N$ & $M$ & $d_{emb}$ & $d_{ff}$ & $d_l$ & $h$ & $S_{batch}$ & $M_{acc}$ & Epochs & Comput. time \\
			\hline
			Autoreg & 2 & 5 & 256 & 1024 & 8 & 16 & 200 & 100 & 6400 & 103.2 \\
			\hline
			\rowcolor{Sandy}
			Birth-death & 1 & 2 & 128 & 1024 & 8 & 8 & 1000 & 100 & 10000 & 11.9 \\
			\hline
			Cascade & 15 & 30 & 128 & 1024 & 8 & 16 & 100 & 100 & 5000 & 128.3 \\
			\hline
			\rowcolor{Sandy}
			Gene express & 2 & 4 & 128 & 1024 & 8 & 8 & 1000 & 100 & 5000 & 195.0 \\
			\hline
			mRNA turnover & 17 & 25 & 128 & 1024 & 8 & 16 & 100 & 100 & 5000 & 122.5 \\
			\hline
			\rowcolor{Sandy}
			Toggle switch & 4 & 8 & 128 & 1024 & 8 & 8 & 500 & 100 & 6000 & 102.5 \\
			\hline
		\end{tabular}
		\begin{tablenotes}
			\small
			\item
			The MET neural network is defined by the embedding dimension $d_{emb}$, the feed-forward neurons $d_{ff}$, the number of decoder layers $d_l$, and the number of attention heads $h$. 
			A scheduled training rate $r_{learn}$ was used, which was optimized to avoid cost function oscillation while maintaining fast convergence as indicated in Method. 
			For each training epoch, $S_{batch}$ state samples were generated for every reward model, and the network weights of $M_{acc}$ models were accumulated for one-step back-propagation. 
			The computational time (in hours) was recorded using a single core GPU of a Tesla-V100 on cloud servers.
		\end{tablenotes}
	\end{threeparttable}
\end{table}

\end{document}